\documentclass[pdflatex,sn-mathphys-num]{sn-jnl}

\usepackage{graphicx}%
\usepackage{multirow}%
\usepackage{amsmath,amssymb,amsfonts}%
\usepackage{amsthm}%
\usepackage{mathrsfs}%
\usepackage[title]{appendix}%
\usepackage{xcolor}%
\usepackage{textcomp}%
\usepackage{manyfoot}%
\usepackage{booktabs}%
\usepackage{algorithm}%
\usepackage{algorithmicx}%
\usepackage{algpseudocode}%
\usepackage{listings}%
\usepackage[utf8]{inputenc}%
\usepackage{comment}

\theoremstyle{thmstyleone}%
\theoremstyle{thmstyletwo}%

\theoremstyle{thmstylethree}%

\begin{document}

\title[Article Title]{Efficient Document Image Dewarping via Hybrid Deep Learning and Cubic Polynomial Geometry Restoration}


\author*[1]{\fnm{Valery} \sur{Istomin}}\email{pc\_valery@mail.ru}

\author[2]{\fnm{Oleg} \sur{Pereziabov}}\email{o.perezyabov@innopolis.ru}
\equalcont{These authors contributed equally to this work.}

\author[2,3]{\fnm{Ilya} \sur{Afanasyev}}\email{i.afanasyev@innopolis.ru}
\equalcont{These authors contributed equally to this work.}

\affil*[1]{
\orgname{Shuya Branch of the Ivanovo State University}, 
\city{Shuya}, \postcode{155908}, 
\country{Russia}}

\affil[2]{\orgdiv{Research Center of the Artificial Intelligence Institute}, 
\orgname{Innopolis University}, 
\city{Innopolis}, \postcode{420500}, 
\country{Russia}}

\affil[3]{
\orgname{Saint Petersburg Electrotechnical University "LETI"}, 
\city{St. Petersburg}, \postcode{197022}, 
\country{Russia}}


\abstract{Camera-captured document images often suffer from geometric distortions caused by paper deformation, perspective distortion, and lens aberrations, significantly reducing OCR accuracy. This study develops an efficient automated method for document image dewarping that balances accuracy with computational efficiency.
We propose a hybrid approach combining deep learning for document detection with classical computer vision for geometry restoration. YOLOv8 performs initial document segmentation and mask generation. Subsequently, classical CV techniques construct a topological 2D grid through cubic polynomial interpolation of document boundaries, followed by image remapping to correct nonlinear distortions. A new annotated dataset and open-source framework are provided to facilitate reproducibility and further research.
Experimental evaluation against state-of-the-art methods (RectiNet, DocGeoNet, DocTr++) and mobile applications (DocScan, CamScanner, TapScanner) demonstrates superior performance. Our method achieves the lowest median Character Error Rate (CER=0.0235), Levenshtein Distance (LD=27.8), and highest Jaro--Winkler similarity (JW=0.902), approaching the quality of scanned originals. The approach requires significantly fewer computational resources and memory compared to pure deep learning solutions while delivering better OCR readability and geometry restoration quality.
The proposed hybrid methodology effectively restores document geometry with computational efficiency superior to existing deep learning approaches, making it suitable for resource-constrained applications while maintaining high-quality document digitization.
Project page: \url{https://github.com/HorizonParadox/DRCCBI}}




\maketitle

\section{Introduction}\label{introduction}

Digital document management systems are increasingly critical across government, business, healthcare, and education sectors. While specialized scanners produce high-quality digitized documents, their availability remains limited, creating demand for efficient camera-based alternatives. Modern smartphones with high-resolution cameras offer convenient document capture, but camera-captured images suffer from multiple degradation factors: perspective distortion from improper camera angles, paper deformations (folds, curves, wrinkles), lens aberrations, unstable lighting conditions, shadows, and motion blur. These distortions significantly impair document readability and drastically reduce Optical Character Recognition (OCR) accuracy, hindering effective integration into electronic document management workflows.

Existing solutions for document image dewarping fall into three categories: classical computer vision approaches using edge detection and geometric transformations, pure deep learning methods that predict dense deformation fields or control points, and hybrid approaches combining both paradigms. Classical methods struggle with complex deformations and varying environmental conditions, while pure deep learning solutions, though achieving promising results on benchmark datasets, demand substantial computational resources (often requiring 15+ GB GPU memory) and training data. Moreover, state-of-the-art deep learning methods (RectiNet, DocGeoNet, DocTr++) frequently exhibit imprecise document boundary delineation, leading to blurred edges that exacerbate rather than correct distortions. Commercial mobile applications (DocScan, CamScanner, TapScanner) demonstrate limited effectiveness with geometric restoration, particularly for documents with severe deformations, often failing to restore original document topology or inadvertently including background artifacts.

This research addresses these limitations by proposing an efficient hybrid methodology that strategically integrates deep learning for robust document detection with classical computer vision for precise geometry restoration. Our approach employs YOLOv8 exclusively for initial document boundary detection and mask generation, a task in which deep learning excels due to its ability to handle complex backgrounds and varying document formats. Subsequently, classical CV techniques construct a topological 2D grid through cubic polynomial interpolation of document boundaries, followed by bicubic image remapping to correct nonlinear distortions. This architecture leverages each paradigm's strengths: deep learning for robust semantic understanding and classical methods for computationally efficient, mathematically precise geometric transformations.

The key advantages of our hybrid approach include: (1) significantly reduced computational requirements compared to pure deep learning solutions, as no specialized GPU training is needed for geometry restoration; (2) superior document boundary precision through polynomial approximation, avoiding the blurred edges characteristic of CNN-based methods; (3) faster inference time with lower memory footprint, enabling deployment on resource-constrained devices; and (4) improved OCR accuracy through topology-preserving geometric correction.

We make three principal contributions:

\textbf{1. Efficient Hybrid Dewarping Methodology:} We propose a novel document geometry restoration approach that combines YOLOv8 deep learning for document outline detection with classical computer vision techniques, specifically cubic polynomial interpolation and bicubic remapping, to create topology-preserving 2D grids that correct nonlinear distortions. This hybrid architecture achieves superior computational efficiency compared to pure deep learning methods while maintaining higher geometric accuracy.

\textbf{2. Open-Source Pipeline and Annotated Dataset:} We developed and released a complete automated document dewarping pipeline (DRCCBI framework) along with an annotated dataset of 392 images featuring diverse document formats and deformation types. The dataset includes documents with varying geometric distortions (folds, curves, perspective distortion), different paper sizes, and challenging environmental conditions. All materials are publicly available at \url{https://github.com/HorizonParadox/DRCCBI} to facilitate reproducibility and enable further research.

\textbf{3. Comprehensive Benchmark Evaluation:} We conducted extensive experimental validation against state-of-the-art deep learning solutions (RectiNet, DocGeoNet, DocTr++) and commercial mobile applications (DocScan, CamScanner, TapScanner). Our method achieves the lowest Character Error Rate (CER=0.0235), Levenshtein Distance (LD=27.8), and highest Jaro-Winkler similarity (JW=0.902) among computational methods, approaching the quality of direct scanning while requiring substantially fewer computational resources. Evaluation using geometry restoration metrics (SSIM, MSE, NRMSE) confirms superior topology reconstruction quality.

The remainder of this paper is organized as follows: Section~\ref{sec:related} reviews related work in document image dewarping, analyzing both deep learning and classical approaches, along with commercial mobile applications. Section~\ref{sec:methodology} presents our hybrid methodology, detailing the YOLOv8-based document detection and the classical CV geometry restoration algorithm. Section~\ref{sec:dataset} describes the dataset and evaluation metrics. Section~\ref{sec:results} presents comparative results against mobile applications and deep learning benchmarks, demonstrating superiority in both OCR-based readability and geometry restoration quality. Section~\ref{sec:conclusion} concludes with discussion and future research directions.

\section{Related Work}\label{sec:related}

Document image dewarping addresses geometric distortions caused by paper deformation, perspective distortion, and lens aberrations in camera-captured images. Existing approaches can be categorized into three paradigms: learning-free methods based on classical computer vision (CV), hybrid methods combining CV and deep learning (DL), and pure deep learning solutions. This section reviews representative methods from each category, with particular focus on state-of-the-art deep learning benchmarks and commercial mobile applications against which we evaluate our approach.

\subsection{Learning-Free and Hybrid Approaches}

Early document dewarping methods relied on classical computer vision techniques. The multistage curvilinear coordinate transform method~\cite{dasgupta2020multistage} employs iterative curvilinear homography with quality estimation based on parallelism, orthogonality, and linearity of text lines. The grid regularization technique~\cite{jiang2022revisiting} minimizes image distortions by enforcing regular grid structures through mathematical optimization. The geometric control points method~\cite{li2023dewarping} uses document boundaries and text lines to create control point grids, applying thin plate spline interpolation for smooth transformation. While these methods demonstrate effectiveness on well-constrained scenarios, they struggle with complex deformations, varying environmental conditions, and require careful parameter tuning.

Hybrid approaches combine deep learning for specific subtasks with classical CV for geometry restoration. The semi-CNN approach~\cite{garai2021dewarping} evaluates pixel position changes using deformation parameters assessed by CNNs on synthetic datasets. Fourier Document Restoration (FDRNet)~\cite{xue2022fourier} employs Fourier transformation and thin-plate splines, using a Coarse and Refinement Transformer to predict control points for dewarping. The text-lines and line segments strategy~\cite{kil2017robust} detects regions of interest, such as horizontal lines for text alignment and document borders, to build a comprehensive understanding of document layout before correction. Despite improved robustness, hybrid methods often still require substantial computational resources for the DL components.

\subsection{Deep Learning Methods}

Pure deep learning approaches dominate recent document dewarping research. DewarpNet~\cite{das2019dewarpnet} utilizes 3D and 2D regression networks for explicit 3D shape modeling and introduced the comprehensive Doc3D dataset with various deformation annotations. DocUNet~\cite{ma2018docunet} employs a U-Net module with intermediate supervision to improve deformation grid prediction accuracy, presenting a broad dataset of geometrically distorted images from mobile devices. The adversarial gated unwarping network~\cite{liu2020geometric} uses pyramid encoder-decoder architecture with gated modules to focus on significant visual features such as text lines and table borders, enhanced through adversarial training.

The ``Learning From Documents in the Wild''~\cite{ma2022learning} method employs two networks: Enet performs edge-based coarse global dewarping using segmentation masks for weakly supervised training, while Tnet refines local deformations learned from document texture. DocReal~\cite{yu2024docreal} offers robust dewarping via an attention-enhanced control point (AECP) module, where Enet performs edge-based dewarping to produce coarse results, and AECP improves precision by predicting control points for local deformations. The displacement flow estimation method~\cite{xie2020dewarping} uses fully convolutional networks to estimate pixel-wise displacements, training on synthetically distorted images to learn correction of various deformation types.

\textbf{State-of-the-Art Benchmarks.} Three methods represent current state-of-the-art performance on standard benchmarks and serve as primary comparisons for our work:

\textbf{RectiNet}~\cite{bandyopadhyay2021gated} introduces a gated and bifurcated stacked U-Net module for document image dewarping. The architecture prevents grid coordinate mixing through bifurcation and uses a gating network to incorporate fine details. Trained on synthetic distorted images and evaluated on real-world datasets, RectiNet demonstrates robust performance on documents with perspective distortions and folds.

\textbf{DocGeoNet}~\cite{feng2022geometric} proposes explicit geometric representation learning by incorporating two key document attributes: 3D shape and text lines. The 3D shape provides global cues for rectifying distorted documents, while text lines offer local geometric features. This dual representation significantly enhances correction quality on documents with complex deformations.

\textbf{DocTr}~\cite{feng2021doctr,feng2023deep} employs dual transformers for geometric dewarping and illumination correction. The first transformer captures global context through self-attention mechanisms and decodes pixel shift solutions for geometric correction. The second transformer removes shadow artifacts post-geometric correction, enhancing visual quality and OCR accuracy. DocTr++ extends this with hierarchical encoder-decoder structure for multi-scale representation, handling unrestricted distorted images including documents with partially missing borders.

\textbf{Common Limitations.} Despite promising results on benchmark datasets, state-of-the-art deep learning methods exhibit several critical limitations. First, they demand substantial computational resources, often requiring 15+ GB GPU memory for training and inference, limiting deployment on resource-constrained devices. Second, these methods frequently demonstrate imprecise document boundary delineation, leading to blurred edges that can exacerbate rather than correct distortions. Third, pure DL approaches require large annotated training datasets and extensive training time. These limitations motivate our hybrid approach that leverages deep learning only for document detection while employing classical CV for efficient, precise geometry restoration.

\subsection{Commercial Mobile Applications}

Commercial mobile document scanning applications represent practical deployments of document dewarping technology. We evaluated three widely-used applications from Google Play Store:

\textbf{DocScan}~\cite{docscan} provides automated document detection and digitization without manual intervention. Analysis revealed that while it handles documents with simple backgrounds effectively, geometric restoration quality degrades significantly with severe paper deformations.

\textbf{CamScanner}~\cite{camscanner} offers document scanning with automatic edge detection and perspective correction. Testing showed limitations in detecting edges when document color is similar to background color or when documents have colored borders. Non-standard sized documents (non-ISO format) pose additional challenges.

\textbf{TapScanner}~\cite{tapscanner} implements automatic document boundary detection with geometry correction capabilities. Evaluation demonstrated that pages of open books are often scanned as single sheets without separation, and documents with geometric distortions frequently retain distorted appearance after processing.

\textbf{Observed Limitations.} Across all evaluated mobile applications, several consistent limitations emerged: (1) false positive detection when fragments of other documents appear in the frame; (2) failure to restore original document topology when significant paper deformations exist; (3) inadvertent inclusion of background artifacts in digitized output; (4) missing portions of documents due to imprecise boundary detection; (5) inability to handle documents with colored borders or non-standard formats. These limitations, combined with the computational constraints of mobile devices, demonstrate the need for efficient hybrid approaches that balance accuracy with resource requirements.

\subsection{Positioning of Our Approach}

The reviewed literature reveals a critical gap: pure deep learning methods achieve high accuracy but demand prohibitive computational resources, while classical CV and mobile applications fail to handle complex deformations effectively. Our hybrid methodology addresses this gap by strategically combining YOLOv8 deep learning, used exclusively for robust document detection (a task requiring semantic understanding), with classical computer vision techniques for geometry restoration via cubic polynomial interpolation and bicubic remapping. This architecture achieves superior computational efficiency compared to pure DL methods (no GPU required for geometry restoration), superior boundary precision through polynomial approximation (avoiding blurred edges characteristic of CNNs), and improved topology preservation compared to mobile applications. The following sections detail our methodology and demonstrate through comprehensive benchmarking that this hybrid approach delivers state-of-the-art OCR accuracy and geometry restoration quality while requiring substantially fewer computational resources than pure deep learning alternatives.

\section{Methodology}\label{sec:methodology}

This section presents our hybrid document image dewarping methodology that strategically integrates deep learning for document detection with classical computer vision for geometry restoration. We first provide an overview of the complete pipeline architecture, then detail each stage: YOLOv8-based document detection with mask generation, followed by the classical CV geometry restoration algorithm employing cubic polynomial interpolation and bicubic remapping.

\subsection{Pipeline Overview}

\begin{figure}[htbp]
\centering
\includegraphics[width=0.8\textwidth]{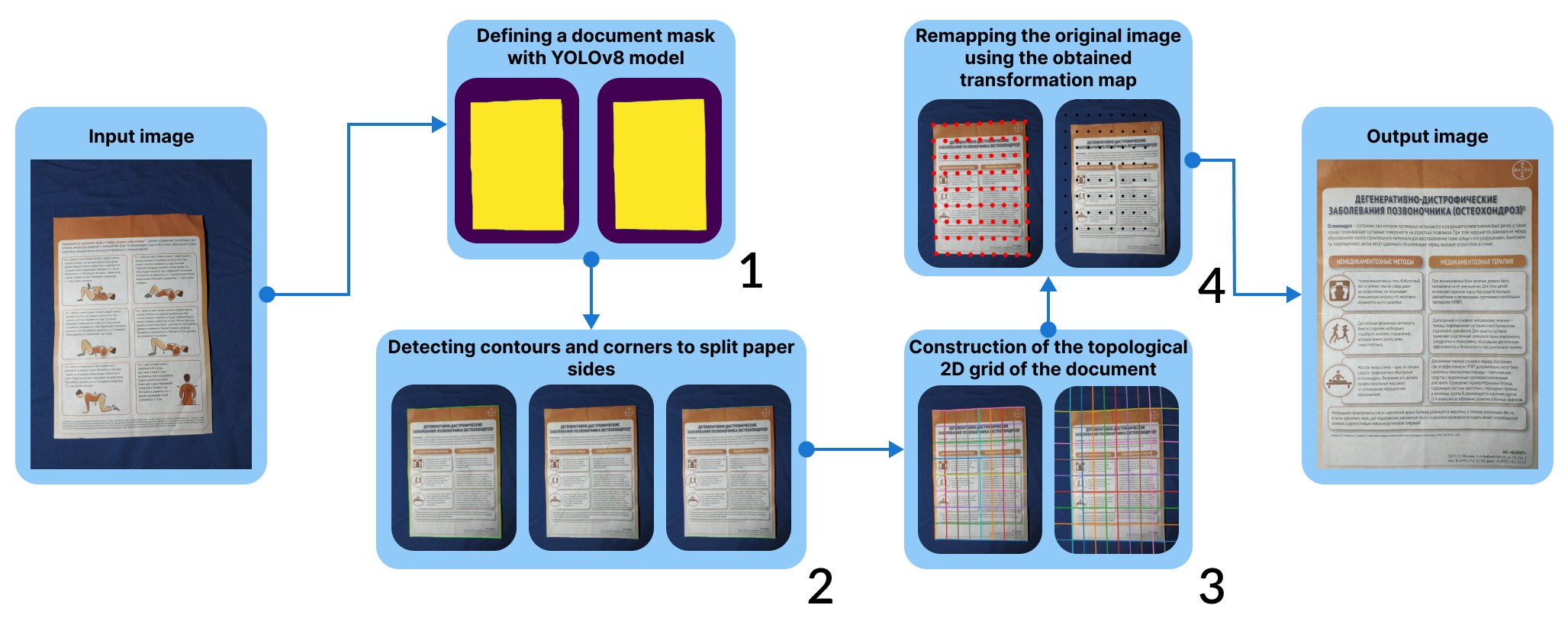}
\caption{Flowchart of the document geometry restoration and dewarping algorithm: 1) Document mask detection with YOLOv8; 2) Contour edge detection and segmentation into document sides; 3) Construction of a topological 2D grid via interpolation with cubic polynomials; 4) Detection of intersection points, grid formation, transformation mapping, and image remapping.}
\label{fig:pipeline}
\end{figure}

Our document dewarping pipeline comprises four principal stages (Figure~\ref{fig:pipeline}).

Let $I = \{(x,y) \mid x \in [0,W), y \in [0,H]\}$ be the set of all pixels in the image of width $W$ and height $H$. Our hybrid document dewarping methodology strategically integrates YOLOv8 deep learning for document detection with classical computer vision for geometry restoration. The complete pipeline comprises four principal stages (Figure~\ref{fig:pipeline}): (1) document mask detection with YOLOv8, (2) contour edge detection and segmentation into document sides, (3) construction of a topological 2D grid via cubic polynomial interpolation, and (4) grid intersection detection, transformation mapping, and image remapping. Algorithm~\ref{alg:dewarping} presents the complete pseudocode.

\begin{algorithm}[t]
\caption{Hybrid Document Image Dewarping}\label{alg:dewarping}
\begin{algorithmic}[1]
\Require Document image $I(x,y)$, dimensions $W \times H$
\Ensure Dewarped image $I'(x,y)$
\State \textbf{Stage 1: Document Detection}
\State $M(x,y) \gets \text{YOLOv8\_Detect}(I(x,y))$
\State $M_{\text{refined}}(x,y) \gets \text{GuidedFilter}(M(x,y), I(x,y))$
\State $M_{\text{binary}}(x,y) \gets \text{OtsuThreshold}(M_{\text{refined}}(x,y))$
\State
\State \textbf{Stage 2: Contour Extraction and Segmentation}
\State $C \gets \text{ExtractContour}(M_{\text{binary}})$
\State $\text{corners} \gets \text{DetectCorners}(C)$
\State $\{S_1, S_2, S_3, S_4\} \gets \text{SegmentSides}(C, \text{corners})$
\State
\State \textbf{Stage 3: Topological Grid Construction}
\For{each side $S_i$}
    \State $L_i(\lambda) \gets \text{LinearInterpolate}(S_i, \lambda)$
    \State $f_i(x) \gets \text{CubicPolynomialFit}(L_i)$
\EndFor
\State $G_{\text{distorted}} \gets \text{ConstructGrid}(\{f_1, f_2, f_3, f_4\})$
\State
\State \textbf{Stage 4: Grid Intersection and Remapping}
\State $G_{\text{intersections}} \gets \text{FindIntersections}(G_{\text{distorted}})$
\State $G_{\text{uniform}} \gets \text{CreateUniformGrid}(W, H)$
\State $\Delta(x,y) \gets \text{ComputeDisplacement}(G_{\text{intersections}}, G_{\text{uniform}})$
\State $I'(x,y) \gets \text{BicubicRemap}(I(x,y), \Delta(x,y))$
\State \Return $I'(x,y)$
\end{algorithmic}
\end{algorithm}

\subsection{Document Mask Detection with YOLOv8}

\textit{A) Document region detection.} YOLOv8~\cite{yolov8_ultralytics} predicts a document region $D(x,y)$ within the input image $I(x,y)$, outputting a probability map:
\begin{equation}
M(x,y) \in [0,1], \quad \forall(x,y) \in I.
\end{equation}
This initial mask provides rough segmentation but may contain noise or imprecise edges. YOLOv8 was trained on a dataset of 392 annotated document images comprising DocUNet~\cite{ma2018docunet}, COCO~\cite{lin2014microsoft}, SmartDoc-QA~\cite{SmartDoc2015}, and custom captures. Training employed the pretrained YOLOv8x-seg model for 150 epochs on Google Colab, achieving superior boundary precision compared to Mask R-CNN alternatives.

\textit{B) Mask refinement using guided filter.} To improve spatial coherence, the guided filter~\cite{he2012guided} is applied:
\begin{equation}
M_{\text{refined}}(x,y) = a(x,y) \cdot D(x,y) + b(x,y),
\end{equation}
where $M_{\text{refined}}(x,y)$ is the refined mask; $a(x,y)$, $b(x,y)$ are the linear regression coefficients, derived from local statistics of initial mask $M(x,y)$ and guidance image $D(x,y)$. The window radius is set to 1\% of the minimum mask dimension, with smoothing parameter $\epsilon$ (half the radius) controlling blur while preserving edge sharpness.

\textit{C) Mask binarization via Otsu's method.} An optimal threshold $t$ minimizes within-class variance~\cite{Otsu1979}:
\begin{equation}
\sigma_\omega^2(t) = \omega_0(t)\sigma_0^2(t) + \omega_1(t)\sigma_1^2(t)
\end{equation}
where $\omega_0(t)$, $\omega_1(t)$ are the probabilities of pixel intensities of the background and foreground classes, respectively; $\sigma_0^2(t)$, $\sigma_1^2(t)$ are the variances within these two classes, resulting in: $M_{\text{binary}}(x,y) = \begin{cases} 1, & \text{if } M_{\text{refined}}(x,y) > t \\ 0, & \text{otherwise} \end{cases}$

\subsection{Document Contour Edge Detection and Segmentation}

\textit{A) Contour extraction.} The largest mask by area is selected, and its contour is identified using edge-following topological structural analysis~\cite{suzuki1985topological}. The contour is approximated using the Douglas-Peucker algorithm~\cite{douglas1973algorithms}, which simplifies geometric shapes while ensuring distances between remaining points remain below specified precision $\epsilon$.

\textit{B) Corner detection and sorting.} The convex hull is computed using the Sklansky algorithm~\cite{sklansky1982finding} with complexity $O(N\log N)$. Corner coordinates are extracted and sorted by calculating the center of mass, then ordering based on increasing angle between the center-to-point direction and positive Y-axis.

\textit{C) Segmentation of document sides.} The document contour is split into four parts: $S = \{S_1, S_2, S_3, S_4\}$, where $S_i \subset P$, and $S_i$ is the set of points belonging to the $i$-th side. Diagonal lines connecting opposite corners partition the document, with edge points identified via cross product calculations and sorted accordingly.

\subsection{Construction of Topological 2D Grid}

\textit{A) Constructing a topological 2D grid.} Opposite sides of the grid are interpolated with curves:
\begin{equation}
L_i(\lambda) = (1-\lambda)P_i + \lambda P_j, \quad \lambda \in [0,1],
\end{equation}
where $L_i$ is the interpolated segment between corresponding points $P_i$, $P_j$, and $\lambda$ is the interpolation parameter. This creates both vertical (left-right) and horizontal (top-bottom) grid lines. Obtained coordinates are smoothed using the Savitzky-Golay filter~\cite{savitzky1964smoothing}.

\textit{B) Curve approximation with cubic polynomials.} Each grid line is fitted with a cubic polynomial via nonlinear least squares:
\begin{equation}
f_i(x) = a_i x^3 + b_i x^2 + c_i x + d_i
\end{equation}
where coefficients $a_i, b_i, c_i, d_i$ are found via least squares optimization. Approximated lines are extrapolated to obtain additional points, increasing point density and ensuring intersection capability between perpendicular grid lines.

\subsection{Grid Intersection, Transformation, and Remapping}

\textit{A) Finding intersection points.} Solving the system of equations: $f_i(x) = g_j(x)$ for each pair of curves, where $g_j(x)$ represents curves formed in the topological 2D grid. The k-d tree algorithm~\cite{bentley1975multidimensional} accelerates nearest neighbor searches with complexity $O(\log N)$, using a threshold of 1\% of the greater document dimension.

\textit{B) Building the final transformation grid.} $G = \{g_{i,j}\}$, where $g_{i,j} = (x_{i,j}, y_{i,j})$, and $G$ is the final grid point set. A uniform rectangular target grid is created with evenly spaced coordinates covering the document dimensions.

\textit{C) Displacement map via cubic interpolation.} The image is transformed using the displacement map:
\begin{equation}
\Delta(x,y) = (\Delta_x(x,y), \Delta_y(x,y)) = (I_x(x,y) - x, I_y(x,y) - y)
\end{equation}
where $I_x(x,y)$ and $I_y(x,y)$ are the x and y components of the displacement field obtained from cubic interpolation between grid points.

\textit{D) Image remapping.} The original image $I(x,y)$ is remapped using the displacement map:
\begin{equation}
I'(x,y) = I(x + \Delta_x(x,y), y + \Delta_y(x,y)).
\end{equation}
This bicubic remapping operation corrects geometric distortions by redistributing pixels according to the transformation from the distorted grid to the uniform rectangular grid, yielding the dewarped output image $I'(x,y)$.

\section{Dataset and Metrics}\label{sec:dataset}

This section details the dataset, evaluation metrics, and implementation specifics used for benchmarking the proposed document image dewarping pipeline.

\subsection{Dataset}

We used a diverse dataset comprising 392 annotated camera-captured document images, made publicly available for reproducibility at \url{https://github.com/HorizonParadox/DRCCBI}. Our dataset integrates:
\begin{itemize}
  \item DocUNet, SmartDoc-QA, and selected images from the COCO ``book'' class;
  \item Additional document photographs collected from various online sources and captured with smartphone cameras;
  \item Manual polygon annotations, converted to the COCO and YOLOv8 formats using Roboflow and MakeSense.
\end{itemize}
The dataset includes documents with various geometric deformations, noise, paper sizes, backgrounds, and lighting conditions. It is split into 359 training, 23 validation, and 10 test images. Full annotation and preprocessing details are documented and versioned on the project GitHub repository.

\subsection{Evaluation Metrics}

Performance was evaluated using both geometry restoration and OCR-based readability metrics:
\begin{itemize}
  \item \textbf{Structural Similarity Index (SSIM)} and \textbf{Normalized Root Mean Square Error (NRMSE)} for geometric fidelity;
  \item \textbf{Mean Squared Error (MSE)} for direct pixel-level comparison;
  \item \textbf{Character Error Rate (CER)}, \textbf{Levenshtein Distance (LD)}, and \textbf{Jaro-Winkler Similarity (JW)} using Tesseract OCR on the dewarped output.
\end{itemize}
All metrics are computed with respect to scanned ground-truth images. OCR-based scores are averaged over all available samples. A strong focus is placed on CER and SSIM, as they are most sensitive to practical restoration performance and end-user document usability.

\section{Experiments and Results}\label{sec:results}

This section presents a comprehensive evaluation of our hybrid document dewarping methodology through comparison with commercial mobile applications and state-of-the-art deep learning solutions. We assess both OCR-based text readability and geometry restoration quality using the metrics defined in Section~\ref{sec:dataset}.

\subsection{Comparison with Mobile Applications}

\begin{figure}[htbp]
\centering
\includegraphics[width=0.7\textwidth]{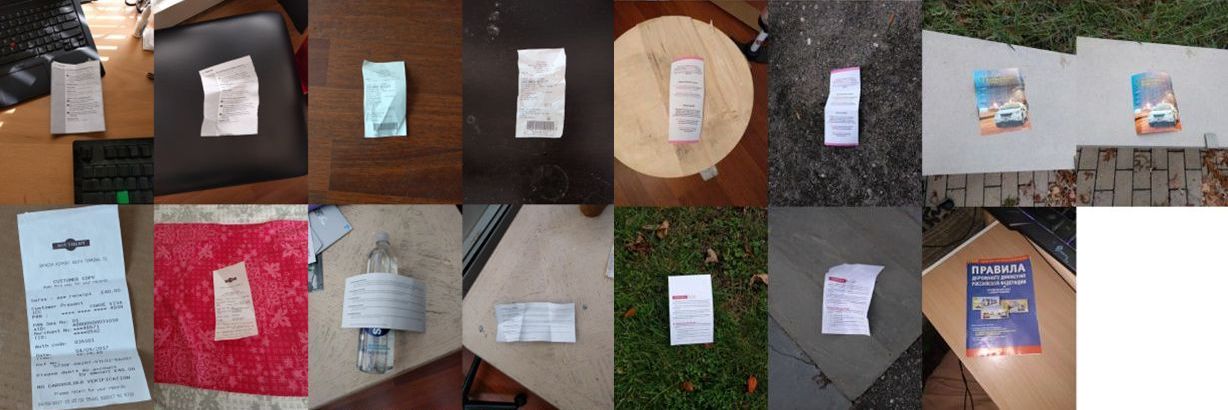}
\caption{Example input images from the test set of 15 camera-captured documents with diverse geometric distortions (e.g., perspective, curvature, folds), used to evaluate mobile scanning applications: DocScan, CamScanner, and TapScanner.}
\label{fig:test_samples}
\end{figure}

We evaluated three widely-used commercial mobile document scanning applications from Google Play Store: DocScan, CamScanner, and TapScanner. For each application, 15 test images with varying deformation types were uploaded without manual adjustments or parameter tuning (Figure~\ref{fig:test_samples}). Two reference images were included for comparison: a scanner-produced digital copy (''Scanned image'') representing the ideal scenario, and the unprocessed camera-captured image (''Original image'') representing the worst-case baseline.

\subsubsection{OCR-Based Text Readability Evaluation}

Text recognition was performed using both EasyOCR~\cite{EasyOCR} and Tesseract OCR~\cite{Tesseract} libraries on all dewarped images. For each document, the original text was manually transcribed to enable accurate metric calculation. Recognized text elements were concatenated into a single string without spaces to facilitate character-level comparison across all evaluation metrics.

\textbf{EasyOCR Results.} Table~\ref{tab:easyocr_mobile} presents median metric values across all 15 test images. Our method achieves the lowest Levenshtein Distance (LD=268), highest Jaro-Winkler Similarity (JW=0.88), and lowest Character Error Rate (CER=0.39) among all computational methods, approaching the quality of direct scanning (LD=50, JW=0.91, CER=0.05). DocScan ranks second with LD=364, demonstrating moderate performance degradation relative to our approach. CamScanner and TapScanner exhibit substantially higher error rates (LD=432 and LD=479, respectively), with CER values exceeding 0.64, indicating significant character recognition failures.

\begin{table}[htbp]
\centering
\caption{OCR-based text readability comparison using EasyOCR library}
\label{tab:easyocr_mobile}
\begin{tabular}{lcccc}
\hline
\textbf{Method} & \textbf{LD $\downarrow$} & \textbf{JW $\uparrow$} & \textbf{CER $\downarrow$} & \textbf{\# Chars} \\
\hline
Original (camera) & 524 & 0.80 & 0.79 & 726 \\
Scanned (ideal) & 50 & 0.91 & 0.05 & 683 \\
\textbf{Our method} & \textbf{268} & \textbf{0.88} & \textbf{0.39} & \textbf{686} \\
DocScan & 364 & 0.82 & 0.47 & 692 \\
CamScanner & 432 & 0.87 & 0.64 & 708 \\
TapScanner & 479 & 0.83 & 0.67 & 706 \\
\hline
\multicolumn{5}{l}{\small True character count: 684} \\
\end{tabular}
\end{table}

The character count analysis reveals that our method produces 686 characters (2 more than ground truth), while mobile applications generate substantially more characters (692-708), indicating spurious character recognition caused by imprecise geometry restoration.

\textbf{Tesseract Results.} Table~\ref{tab:tesseract_mobile} presents Tesseract OCR evaluation results, which include recognition confidence scores. Our method maintains first rank with LD=145, JW=0.82, and CER=0.21, achieving 96\% recognition confidence-matching the scanned image baseline. DocScan follows with LD=160 but substantially lower confidence (88\%). TapScanner demonstrates improved performance relative to EasyOCR results (LD=175), outperforming CamScanner (LD=298). The character count from our method (679) most closely approximates ground truth among all computational solutions.

\begin{table}[htbp]
\centering
\caption{OCR-based text readability comparison using Tesseract library}
\label{tab:tesseract_mobile}
\begin{tabular}{lccccc}
\hline
\textbf{Method} & \textbf{LD $\downarrow$} & \textbf{JW $\uparrow$} & \textbf{CER $\downarrow$} & \textbf{Conf. $\uparrow$} & \textbf{\# Chars} \\
\hline
Original (camera) & 879 & 0.67 & 1.32 & 31 & 1150 \\
Scanned (ideal) & 22 & 0.90 & 0.03 & 96 & 688 \\
\textbf{Our method} & \textbf{145} & \textbf{0.82} & \textbf{0.21} & \textbf{96} & \textbf{679} \\
DocScan & 160 & 0.80 & 0.22 & 88 & 641 \\
CamScanner & 298 & 0.75 & 0.43 & 85 & 820 \\
TapScanner & 175 & 0.79 & 0.26 & 89 & 673 \\
\hline
\multicolumn{6}{l}{\small True character count: 684} \\
\end{tabular}
\end{table}

\subsubsection{Geometry Restoration Quality}

Document topology reconstruction quality was assessed using SSIM, MSE, and NRMSE metrics, with the scanned image serving as ground truth reference. Table~\ref{tab:geometry_mobile} presents median values across all test images.

\begin{table}[htbp]
\centering
\caption{Geometry restoration quality relative to scanned ground truth}
\label{tab:geometry_mobile}
\begin{tabular}{lccc}
\hline
\textbf{Method} & \textbf{SSIM $\uparrow$} & \textbf{MSE $\downarrow$} & \textbf{NRMSE $\downarrow$} \\
\hline
Original (camera) & 0.27 & 72542 & 0.63 \\
\textbf{Our method} & \textbf{0.66} & \textbf{15476} & \textbf{0.29} \\
DocScan & 0.53 & 16528 & 0.30 \\
CamScanner & 0.49 & 20779 & 0.40 \\
TapScanner & 0.61 & 18828 & 0.32 \\
\hline
\end{tabular}
\end{table}

Our method achieves the highest SSIM (0.66) and lowest error metrics (MSE=15476, NRMSE=0.29), demonstrating superior geometric fidelity. TapScanner ranks second in SSIM (0.61) but exhibits higher MSE, while DocScan shows competitive NRMSE (0.30) despite lower structural similarity. CamScanner demonstrates the weakest geometry restoration among evaluated applications.

The substantial improvement over the camera-captured baseline (SSIM improvement from 0.27 to 0.66) validates the effectiveness of our topology-preserving grid construction approach. The relatively small gap between our method and DocScan in NRMSE (0.29 vs. 0.30) contrasts sharply with the larger differences in OCR readability metrics, suggesting that precise boundary delineation and topology preservation-strengths of our cubic polynomial interpolation approach-contribute more significantly to OCR performance than overall geometric similarity alone.

\subsection{Comparison with Deep Learning Methods}

We conducted comprehensive benchmarking against three state-of-the-art deep learning document dewarping methods: RectiNet, DocGeoNet, and DocTr++. These models represent current best practices in pure deep learning approaches and demonstrate strong performance on standard benchmarks such as DocUNet dataset.

\subsubsection{Visual Quality Assessment}

\begin{figure}[htbp]
\centering
\includegraphics[width=0.5\textwidth]{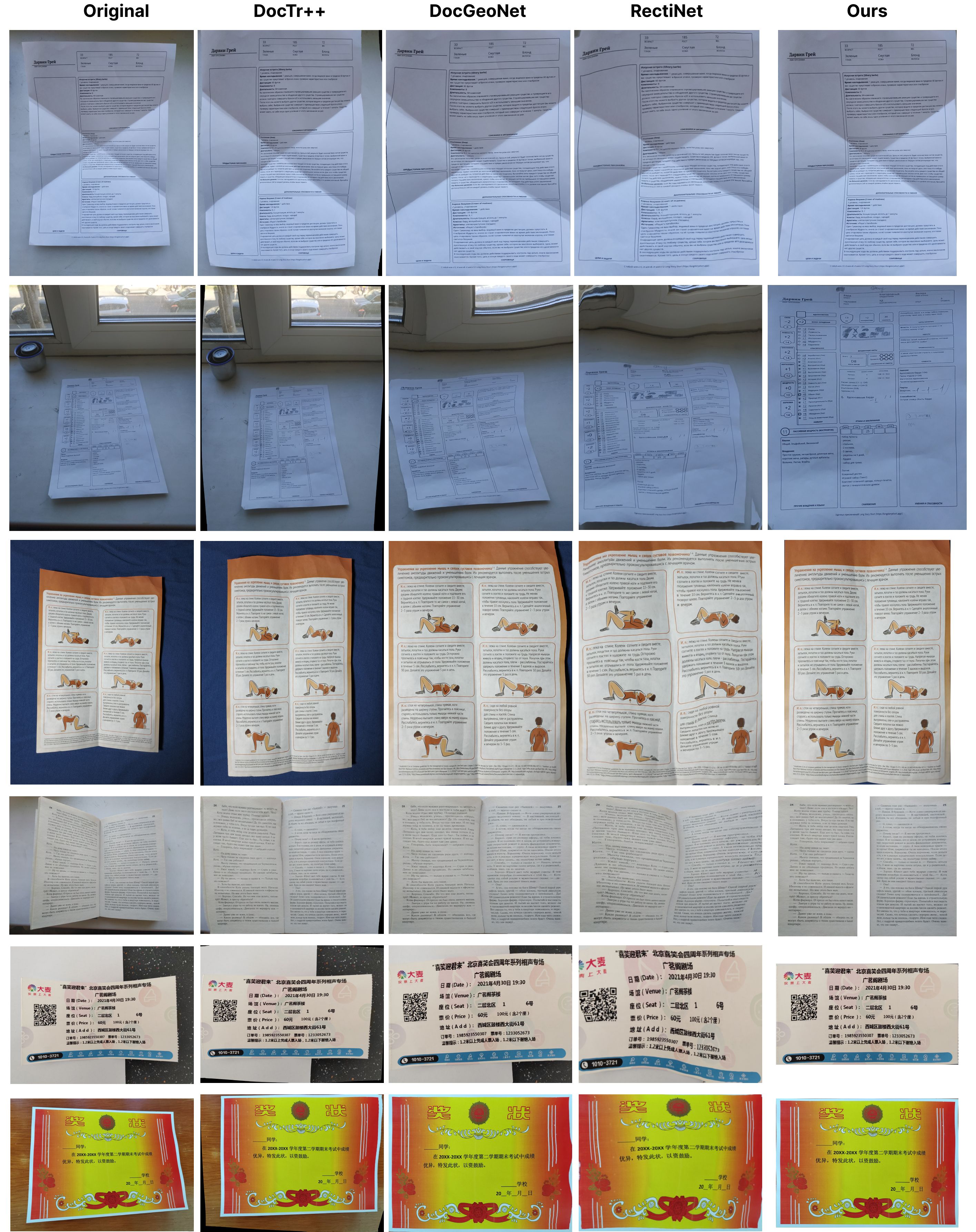}
\caption{The visual comparative analysis of documents reconstructed by our algorithm and popular desktop DL models - DocTr++,
DocGeoNet and RectiNet.}
\label{fig:dl_comparison_1}
\end{figure}

\begin{figure}[htbp]
\centering
\includegraphics[width=0.5\textwidth]{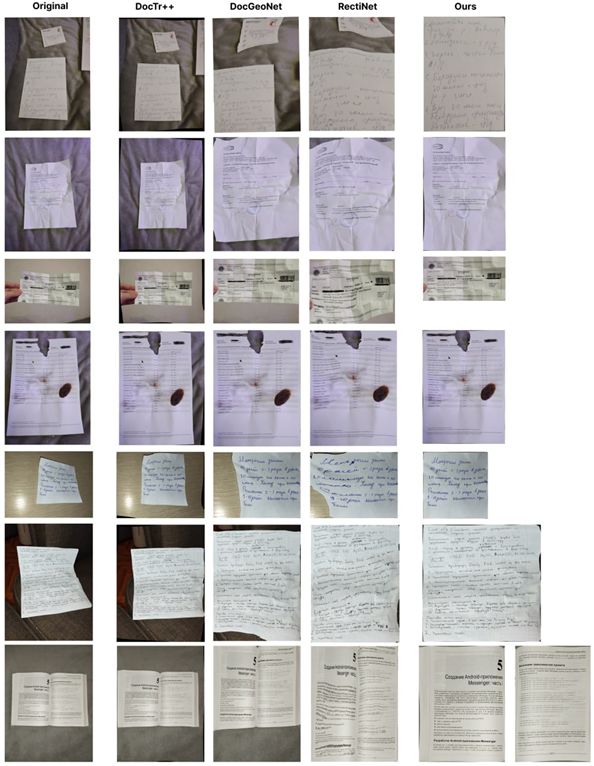}
\caption{Visual comparative analysis of documents reconstructed by the proposed algorithm and popular deep learning models (DocTr++, DocGeoNet, RectiNet) on an additional dataset. Multiple documents with diverse geometric distortions, paper colors, and lighting conditions demonstrate consistent superior boundary precision and topology preservation of our method, resulting in sharper text edges and minimal spurious artifacts compared to all alternatives.}
\label{fig:dl_comparison_2}
\end{figure}

Figures~\ref{fig:dl_comparison_1} and~\ref{fig:dl_comparison_2} present visual comparison of geometry restoration results across multiple documents with varying deformation severities, formats, and color palettes. Our method consistently produces sharper boundary delineation compared to DL alternatives, which exhibit characteristic blurred edges-a common artifact of CNN-based pixel displacement prediction.

In the first example (top row), all methods encounter difficulty with the curved left boundary due to severe physical paper deformation. However, our approach produces the most presentable result with minimal boundary waviness. RectiNet and DocGeoNet introduce substantial geometric artifacts, while DocTr++ partially preserves document structure but with noticeable edge blur. This pattern repeats across all test cases: our classical CV geometry restoration maintains crisp boundaries through explicit polynomial approximation, whereas neural network methods struggle to precisely delineate document edges despite their sophisticated architectures.

\subsubsection{OCR-Based Text Readability}

Tables~\ref{tab:easyocr_dl} and~\ref{tab:tesseract_dl} present OCR-based readability metrics for documents processed by our method and DL benchmarks, using the same 15 test images as the mobile application comparison (Figure~\ref{fig:test_samples}).

\begin{table}[htbp]
\centering
\caption{Text readability comparison with DL methods using EasyOCR}
\label{tab:easyocr_dl}
\begin{tabular}{lcccc}
\hline
\textbf{Method} & \textbf{LD $\downarrow$} & \textbf{JW $\uparrow$} & \textbf{CER $\downarrow$} & \textbf{\# Chars} \\
\hline
Original (camera) & 361 & 0.80 & 0.79 & 673 \\
Scanned (ideal) & 20 & 0.89 & 0.05 & 683 \\
\textbf{Our method} & \textbf{126} & \textbf{0.87} & \textbf{0.33} & \textbf{674} \\
RectiNet & 465 & 0.78 & 0.80 & 680 \\
DocGeoNet & 287 & 0.83 & 0.70 & 685 \\
DocTr++ & 165 & 0.81 & 0.51 & 668 \\
\hline
\multicolumn{5}{l}{\small True character count: 684} \\
\end{tabular}
\end{table}

\textbf{EasyOCR Results.} Our method substantially outperforms all DL alternatives across all metrics (Table~\ref{tab:easyocr_dl}). With LD=126 and CER=0.33, our approach achieves 3.7$\times$ lower error rate than RectiNet (CER=0.80) and 2.1$\times$ lower than DocGeoNet (CER=0.70). Even DocTr++, the strongest DL baseline, exhibits 54\% higher CER (0.51 vs. 0.33). Character count accuracy further validates our method's superiority: 674 characters (10 fewer than ground truth) compared to DocTr++'s 668, suggesting our approach introduces minimal spurious recognitions.

\textbf{Tesseract Results.} Tesseract OCR evaluation (Table~\ref{tab:tesseract_dl}) reinforces our method's dominance. Recognition confidence (an internal Tesseract metric) reaches 84\% for our approach, dramatically exceeding all DL methods, which score only 33-37\%. This 2.3-2.5$\times$ confidence advantage indicates that our geometry restoration produces substantially more recognizable character shapes. RectiNet performs poorly (LD=904, CER=1.60), generating 2545 characters, i.e. 3.7$\times$ more than ground truth, indicating catastrophic over-segmentation from blurred boundaries.

The character count metric reveals a critical insight: while our method underestimates slightly (554 vs. 684 true characters), DL methods dramatically overestimate (939-2545 characters), suggesting they create spurious text regions from geometric artifacts. This pattern explains why high character counts in Table~\ref{tab:easyocr_dl} do not correlate with recognition quality, as the excess characters are predominantly false positives.

\begin{table}[htbp]
\centering
\caption{Text readability comparison with DL methods using Tesseract}
\label{tab:tesseract_dl}
\begin{tabular}{lccccc}
\hline
\textbf{Method} & \textbf{LD $\downarrow$} & \textbf{JW $\uparrow$} & \textbf{CER $\downarrow$} & \textbf{Conf. $\uparrow$} & \textbf{\# Chars} \\
\hline
Original (camera) & 974 & 0.60 & 1.32 & 31 & 1042 \\
Scanned (ideal) & 25 & 0.90 & 0.04 & 96 & 688 \\
\textbf{Our method} & \textbf{194} & \textbf{0.72} & \textbf{0.43} & \textbf{84} & \textbf{554} \\
RectiNet & 904 & 0.57 & 1.60 & 33 & 2545 \\
DocGeoNet & 620 & 0.60 & 0.94 & 36 & 1123 \\
DocTr++ & 488 & 0.68 & 0.79 & 37 & 939 \\
\hline
\multicolumn{6}{l}{\small True character count: 684} \\
\end{tabular}
\end{table}

\subsubsection{OCR Enhancement with DeepSeek-VL}

Beyond raw OCR metrics, we evaluated post-processing using DeepSeek-VL multimodal language model for semantic text correction on an additional diverse document dataset. This approach leverages the model's understanding of document layout and language patterns to correct systematic OCR errors beyond character-level recognition. Figure~\ref{fig:test_samples} presents the varied document dataset used for DeepSeek-VL evaluation, demonstrating different deformation types, formats, and color palettes.

Table~\ref{tab:deepseek_correction} presents comparative OCR results where all methods (ours, RectiNet, DocGeoNet, DocTr++, DocScan) first dewarp documents using their respective algorithms, then OCR is performed using DeepSeek-VL on the dewarped outputs. Our method achieves the lowest error metrics: LD=27.8, JW=0.902, and CER=0.0235, approaching scanned reference quality (LD=22.5, CER=0.0675). This substantially outperforms all alternatives: DocScan (LD=41.5, CER=0.0515), DocGeoNet (LD=45.5, CER=0.077), RectiNet (LD=130.5, CER=0.3), and DocTr++ (LD=175, CER=0.2655).

\begin{table}[htbp]
\centering
\caption{Document geometry reconstruction comparison with DeepSeek-VL OCR}
\label{tab:deepseek_correction}
\begin{tabular}{lccc}
\hline
\textbf{Method} & \textbf{LD $\downarrow$} & \textbf{JW $\uparrow$} & \textbf{CER $\downarrow$} \\
\hline
Original (camera) & 167.8 & 0.856 & 0.194 \\
Scanned (ideal) & 22.5 & 0.9075 & 0.0675 \\
\textbf{Our method} & \textbf{27.8} & \textbf{0.902} & \textbf{0.0235} \\
RectiNet & 130.5 & 0.831 & 0.3 \\
DocGeoNet & 45.5 & 0.879 & 0.077 \\
DocTr++ & 175 & 0.8525 & 0.2655 \\
DocScan & 41.5 & 0.901 & 0.0515 \\
\hline
\end{tabular}
\end{table}

The results validate that superior geometric restoration directly translates to improved OCR performance even with advanced post-processing. While DeepSeek-VL's multimodal understanding provides sophisticated text correction, the fundamental quality of geometry restoration remains the primary determinant of OCR accuracy. Our method's explicit topology preservation through cubic polynomial grid construction produces text regions with minimal distortion, enabling DeepSeek-VL to achieve near-scanned quality results (CER=0.0235 vs. scanned CER=0.0675).

Notably, DocScan, which ranks second among computational methods, exhibits a CER that is 2.2 times higher than that of our approach, while pure DL methods demonstrate substantially degraded performance: DocTr++ achieves CER=0.2655 (11.3$\times$ worse than ours) despite its sophisticated transformer architecture. This dramatic performance gap confirms that learned pixel displacement fields struggle to preserve the geometric constraints essential for accurate text recognition, whereas our analytical polynomial-based approach mathematically guarantees topology preservation.

\subsubsection{Geometry Restoration Quality}

Our method achieves the highest SSIM (0.56) and lowest errors (MSE=4527, NRMSE=0.27), outperforming all DL baselines by substantial margins. RectiNet, the second-best performer, exhibits SSIM=0.29 (48\% lower) and MSE=15499 (3.4$\times$ higher). DocGeoNet and DocTr++, despite their sophisticated geometric modeling and transformer architectures, perform barely better than the uncorrected camera baseline, with SSIM values of 0.19 and 0.16, respectively.

These results validate our hypothesis that explicit topology modeling through cubic polynomial grid construction preserves document structure more effectively than learned pixel displacement fields. DL methods optimize for pixel-wise similarity on training data but struggle to generalize to the geometric constraints of document topology, particularly boundary preservation.

Table~\ref{tab:geometry_dl} quantifies geometric fidelity using SSIM, MSE, and NRMSE metrics.

\begin{table}[htbp]
\centering
\caption{Geometry restoration quality comparison with DL methods}
\label{tab:geometry_dl}
\begin{tabular}{lccc}
\hline
\textbf{Method} & \textbf{SSIM $\uparrow$} & \textbf{MSE $\downarrow$} & \textbf{NRMSE $\downarrow$} \\
\hline
Original (camera) & 0.18 & 24180 & 0.63 \\
\textbf{Our method} & \textbf{0.56} & \textbf{4527} & \textbf{0.27} \\
RectiNet & 0.29 & 15499 & 0.50 \\
DocGeoNet & 0.19 & 24470 & 0.61 \\
DocTr++ & 0.16 & 25840 & 0.67 \\
\hline
\end{tabular}
\end{table}

\subsection{Computational Efficiency Analysis}

A critical advantage of our hybrid approach is computational efficiency. The complete pipeline is implemented in Python using the OpenCV library for classical computer vision operations, the \texttt{Ultralytics} library for YOLOv8 integration \cite{yolov8_ultralytics}, and NumPy for numerical computations. The YOLOv8 model was trained in Google Colab using an NVIDIA T4 GPU, and the resulting weights were deployed for inference on a local laptop equipped with an Intel Core i5-10300H CPU, 16~GB of RAM, and an NVIDIA GTX 1650 Ti GPU (4~GB VRAM). Performance was evaluated across 11 diverse test documents from our dataset (resolution $4080 \times 3072$ pixels, JPEG format, average file size $\sim$5~MB). The complete pipeline (YOLOv8 detection + geometry restoration) achieves the performance shown in Table~\ref{tab:timing}.

\begin{table}[htbp]
\centering
\caption{Processing time per image (seconds)}
\label{tab:timing}
\begin{tabular}{lcccc}
\hline
\textbf{Metric} & \textbf{Mean} & \textbf{Median} & \textbf{Minimum} & \textbf{Maximum} \\
\hline
Time (s) & 5.12 & 4.97 & 3.79 & 7.64 \\
\hline
\end{tabular}
\end{table}

Notably, the geometry restoration stage, which comprises cubic polynomial fitting and bicubic remapping, runs entirely on the CPU and consumes less than 2~GB of RAM. This modular design allows independent optimisation of the detection and restoration stages and makes our solution highly suitable for deployment in resource-constrained environments where GPU acceleration is either unavailable or economically impractical. In contrast, pure deep learning approaches for comparable dewarping tasks typically require 15+~GB of GPU memory, rendering them infeasible on edge or mobile hardware.

While we did not benchmark pure deep learning methods (RectiNet, DocGeoNet, DocTr++) on our hardware due to environment configuration challenges, published literature provides useful context. DocGeoNet~\cite{feng2022geometric}, a transformer-based method with 24.8 million parameters, was evaluated on an RTX 2080Ti GPU for 1080p images. DocTr++~\cite{feng2023deep} achieves 11.29 FPS with 26.6 million parameters, demonstrating high computational efficiency among deep learning approaches.

\section{Conclusion and Discussion}\label{sec:conclusion}

Our comprehensive evaluation on the proposed document images dataset confirms that the hybrid methodology, which combines deep learning for semantic boundary detection with classical computer vision for geometry restoration, consistently outperforms both commercial mobile applications and state-of-the-art deep learning approaches (including RectiNet, DocGeoNet, and DocTr++) in terms of geometric fidelity, OCR accuracy, and computational efficiency. 

The key strengths of our approach stem from three design principles:

\begin{enumerate}
    \item \textbf{Explicit topology preservation}: By constructing a 2D grid via cubic polynomial interpolation of opposite document sides, we enforce geometric constraints (e.g., parallelism of opposite edges) that are often violated by end-to-end neural models learning deformations implicitly from data.
    
    \item \textbf{Boundary precision}: Analytical fitting of document contours yields sharp, well-defined edges, which are critical for downstream OCR, whereas dense displacement fields predicted by neural networks inherently suffer from spatial smoothing.
    
    \item \textbf{Resource-aware architecture}: Deep learning is confined to the detection stage (where semantic understanding is essential), while geometry restoration leverages lightweight classical algorithms that run entirely on CPU and require less than 2~GB of RAM. This enables deployment on smartphones and edge devices, where pure deep learning solutions (typically demanding 15-22~GB GPU memory) are infeasible.
\end{enumerate}

\noindent\textbf{Summary of contributions.}  
This work makes three principal contributions to the field of document image analysis:
\begin{enumerate}
    \item A novel hybrid dewarping pipeline that strategically integrates YOLOv8 for robust boundary detection with classical CV for efficient geometry restoration via cubic polynomial interpolation and bicubic remapping;
    \item An open-source implementation and a curated dataset of 392 real-world camera-captured document images, annotated with diverse deformations, lighting conditions, and backgrounds, addressing the scarcity of publicly available dewarping benchmarks under realistic capture settings;
    \item A comprehensive experimental evaluation against both deep learning baselines and commercial tools, demonstrating superior performance in OCR metrics (CER, Jaro-Winkler similarity) and geometric quality (SSIM, MSE, NRMSE).
\end{enumerate}

Despite its advantages, the method has limitations. Performance degrades both under extreme occlusions, where polynomial fitting becomes ill-conditioned, and on non-quadrilateral documents, such as circular or irregularly shaped certificates, highlighting the current reliance on 2D contour-based modeling.

Future work will explore: (i) integration of depth sensors (e.g., LiDAR) for 3D-aware dewarping of severely curved surfaces; (ii) adoption of advanced detection architectures (e.g., YOLOv11 with GELAN); (iii) extension to multi-page layouts and real-time mobile deployment via TensorFlow Lite or ONNX Runtime; and (iv) domain adaptation techniques for handling degraded archival or multilingual documents.

In summary, our results show that the most viable path toward practical, high-quality, and deployable document digitization systems lies in a principled fusion of deep learning and classical computer vision, not in end-to-end neural architectures. The full pipeline, dataset, and evaluation code are publicly available at \url{https://github.com/HorizonParadox/DRCCBI}.

\backmatter

\end{document}